\documentclass[conference]{IEEEtran}
\IEEEoverridecommandlockouts
\usepackage{cite}
\usepackage{amsmath,amssymb,amsfonts}
\usepackage{algorithmic}
\usepackage{graphicx}
\usepackage{textcomp}
\usepackage{xcolor}
\usepackage{listings}
\usepackage{url}
\usepackage{array}
\usepackage{tikz} 
\usepackage{balance}

\def\BibTeX{{\rm B\kern-.05em{\sc i\kern-.025em b}\kern-.08em
    T\kern-.1667em\lower.7ex\hbox{E}\kern-.125emX}}

\newcolumntype{P}[1]{>{\centering\arraybackslash}p{#1}}
    
\begin{document}

\title{An Iterative Approach to Topic Modelling\\
\thanks{We acknowledge the support of the Student Work on Campus program as well as the Statistics and Data Analytics Department at Langara College.}
}

\makeatletter
\newcommand{\linebreakand}{%
 \end{@IEEEauthorhalign}
 \hfill\mbox{}\par
 \mbox{}\hfill\begin{@IEEEauthorhalign}
}
\makeatother

\author{
\IEEEauthorblockN{Albert Wong}
\IEEEauthorblockA{\textit{Statistics and Data Analytics} \\
\textit{Langara College}\\
Vancouver, Canada }

\and

\IEEEauthorblockN{Florence Wing Yau Cheng}
\IEEEauthorblockA{\textit{Statistics and Data Analytics} \\
\textit{Langara College}\\
Vancouver, Canada }

\and
\IEEEauthorblockN{Ashley Keung}
\IEEEauthorblockA{\textit{Statistics and Data Analytics} \\
\textit{Langara College}\\
Vancouver, Canada }

\linebreakand 

\and
\IEEEauthorblockN{Yamileth Hercules}
\IEEEauthorblockA{\textit{Statistics and Data Analytics} \\
\textit{Langara College}\\
Vancouver, Canada}

\and
\IEEEauthorblockN{Mary Alexandra Garcia}
\IEEEauthorblockA{\textit{Statistics and Data Analytics} \\
\textit{Langara College}\\
Vancouver, Canada}

\and
\IEEEauthorblockN{Yew-Wei Lim}
\IEEEauthorblockA{\textit{Statistics and Data Analytics} \\
\textit{Langara College}\\
Vancouver, Canada }

\linebreakand 

\IEEEauthorblockN{Lien Pham}
\IEEEauthorblockA{\textit{Statistics and Data Analytics} \\
\textit{Langara College}\\
Vancouver, Canada}

}

\maketitle

\begin{abstract}
Topic modelling has become increasingly popular for summarizing text data, such as social media posts and articles. However, topic modelling is usually completed in ``one shot". Assessing the quality of resulting topics is challenging. No effective methods or measures have been developed for assessing the results or for making further enhancements to the topics. In this research, we propose to use an iterative process to perform topic modelling that gives rise to a sense of completeness of the resulting topics when the process is complete. Using the BERTopic package, a popular method in topic modelling, we demonstrate how the modelling process can be applied iteratively to arrive at a set of topics that could not be further improved upon using one of the three selected measures for clustering comparison as the decision criteria. This demonstration is conducted using a subset of the COVIDSenti-A dataset. The early success leads us to believe that further research using in using this approach in conjunction with other topic modelling algorithms could be viable.
\end{abstract}


\section{Introduction}
With the popularity of social media, many people are using it as a platform for expressing their opinions through postings and tweets. From a research perspective, it is challenging to extract useful information from this kind of content. With the help of natural language processing and machine learning algorithms, we are beginning to mine the information embedded in these texts. One such approach is topic modelling, a set of unsupervised learning techniques that aim to classify a set of texts into several ``topics" based on the statistical similarity of the content of these texts.

Topic modelling has been around for the last fifteen years. A review of research work in the recent years can be found in the Literature Review section below. Most, if not all, of the techniques used for topic modelling use a "one-shot" approach, that is, the technique employed gives the researcher a set of topics (clusters of the texts) by going through the data once. Whether the result of this clustering is good is not easy to determine. As well, there is no established process to ``fine tune" the results.

In this research project, we employ the BERTopic package \cite{Grootendorst2022BERTopic:Procedure}, one of the standardized approaches in topic modelling, to demonstrate the viability of using an iterative approach. A data set is used with this package to show how an iterative approach can be used to obtain a ``final" set of topics with the assurance that no further ``improvements" need to be made using the BERTopic approach. 

One of the key ideas in the iterative approach is to compare clustering results in successive steps of the iteration. As explained further in the Literature Review section, there are quite a few measures that have been developed over the years in comparing clusterings. There are also libraries in R and Python for the calculation of these measures. For this research, we will use the typical one from each of the three classes of measures for evaluation: the Modified Rand Index, the Van Dongen measure, and the normalized variation of information index.

Our main contribution is therefore the illustration of the validity and potential of using this iterative approach in topic modelling. A review of the relevant literature is presented in Section II. Section III gives details of the data set used to demonstrate the iterative approach. The topic modelling process and the iterative steps are described respectively in Sections IV and V. Main results of the research are presented in Section VI. Finally, discussions around future work and conclusions can be found respectively in Sections VII and VIII.

\section{Literature Review}

Twitter data has been extensively examined to unearth information that individuals share on this platform. This information needs to be revealed through the utilization of machine learning algorithms such as topic modelling. The practical application of topic modelling is exemplified in the study conducted by Wicke and Bolognesi \cite{Wicke2020FramingTwitter}, where they scrutinized COVID-19 discourse on Twitter. Through the application of a topic modelling technique, they successfully identified prevailing topics and framed patterns within the discussions. Likewise, Curiskis, Drake, Osborn, and Kennedy \cite{Curiskis2020AnReddit} embarked on an exploration of document clustering and topic modelling within online social networks. Their research accentuated the necessity for innovative methodologies to address the challenges posed by the noisy nature of social media data and underscored the limitations of traditional techniques.

Topic modelling conventionally draws from vector space representations, with bag-of-words techniques standing as a prime example. One well-established avenue involves the utilization of a pivotal feature matrix, known as the term-frequency inverse document frequency matrix (TF-IDF matrix). The objective is to pinpoint the most effective methodology among a spectrum of text embedding techniques. These techniques encompass count-based strategies rooted in the bag-of-words framework and prediction-based methods that harness the power of sequential word patterns. \cite{Vayansky2020AMethods}

Extensive research has consistently underlined the effectiveness of incorporating semantic text similarity into document analysis, irrespective of its length. In a comparative exploration undertaken by Shahmirzadi, Lugowski, and Younge \cite{Shahmirzadi2018TextStudy}, the potential of semantic text similarity takes centre stage, especially in gauging the similarity between patents within a voluminous corpus. Their investigation delves deeply into evaluating Vector Space Models (VSMs), closely examining the performance of diverse text embedding strategies. The study underscored the practical prowess of the straightforward TF-IDF model, outshining more intricate alternatives such as Latent Semantic Indexing (LSI) and Document-to-Vector (D2V). This discovery resonates with our approach, where the incorporation of contextual embeddings like BERT is set to amplify the efficacy of topic modelling, further emphasizing the pivotal importance of judicious method selection.

Similarly, Qiang, Qian, Li, Yuan, and Wu \cite{Qiang2019ShortSurvey} ventured into the realm of short text topic modelling, acknowledging the challenges presented by limited word co-occurrence information. By categorizing methods into distinct groups and examining their performance, their research aligned seamlessly with our focus on iteratively refining topic modelling through enhancements. 

In the pursuit of selecting the most suitable embedding technique, Thompson and Mimno's exploration \cite{Thompson2020TopicClusters} of topic modelling with contextualized word representation clusters showcased the supremacy of token-level contextual embeddings, exemplified by BERT. Through their investigation of BERT and GPT2 embeddings coupled with k-means clustering, their research demonstrated the potential of these embeddings in generating coherent topics.

Latent Dirichlet Allocation (LDA) has emerged as a foundational technique for revealing latent themes in various textual data sources. Negara, Triadi, and Andryani \cite{Negara2019TopicMethod} investigated Twitter data topics using LDA, highlighting its effectiveness in generating weighted topic lists for documents. However, they recognized limitations with concise text datasets and introduced Twitter-LDA for tailored results. Similarly, Jelodar et al. \cite{Jelodar2019LatentSurvey} presented an in-depth survey on LDA's widespread use in uncovering latent structures within textual data across domains and social networks.

An alternative approach is seen in Angelov's introduction \cite{Angelov2020Top2Vec:Topics} of the Top2Vec method, leveraging neural network-based topic representations. Unlike LDA, Top2Vec captures both syntactic and semantic word relationships, dynamically determines topic numbers, and can outperform LDA in identifying informative topics. These studies collectively contribute to the evolution of topic modelling techniques for varied applications, including social media analysis.

Combining cluster analysis with advanced techniques like BERT enhances the topic modelling landscape. George and Sumathy \cite{George2023AnModeling} propose an integrated framework of BERT, LDA, and clustering algorithms for improved topic modelling. This hybrid model utilizes dimension reduction (PCA, t-SNE, UMAP) and K-means clustering to generate reduced-dimension topic clusters. Similarly, Ogunleye et al. \cite{Ogunleye2023ComparisonContext} compare topic modelling methods in banking using LDA, LSI, Hierarchical Dirichlet Process (HDP), and Bertopic. BERT's synergy with Kernel PCA and K-means achieves the highest coherence score. Their approach involves BERT, UMAP, and clustering to uncover nuanced themes for richer insights within the text corpus.

Some forms of an iterative approach have enhanced clustering algorithms, boosting accuracy and performance. Ismkhan \cite{Ismkhan2018AnK-means} introduces an iterative version of K-means clustering by removing and dividing clusters to enhance accuracy. Ding et al \cite{Ding2018IterClust:Analysis} present iterClust, a method that iterates over various clustering methods for subcluster analysis. Lin et al. \cite{Lin2022AAlgorithm} propose a general iterative clustering (GIC) algorithm that iterates between supervised classifiers and the Random Forest algorithm, enhancing clustering performance.

Numerous studies have endeavoured to devise methodologies for effectively comparing clustering or topic modelling outcomes. Meilǎ \cite{Meila2005ComparingView} introduces an innovative approach that centred on the "variation of information" (VI) concept, probing the measurement of information disparity between different clusterings. Rezaei and Fränti \cite{Rezaei2016SetValidity} delve into external cluster validity measures, accentuating the importance of set-matching approaches. Emphasizing the necessity of accounting for chance within information-theoretic clustering measures, Vinh, Epps, and Bailey \cite{XuanVinh2010InformationChance} also contribute to the discourse.

The comparison of clustering solutions stands as a crucial facet of evaluation. Vinh, Epps, and Bailey \cite{XuanVinh2010InformationChance} delve into information-theoretic measures, specifically highlighting the suitability of NID and NVI. Meila \cite{Meila2007ComparingDistance} introduces the Variation of Information (VI) as a mechanism to gauge differences in information acquisition and loss among clusterings. Hubert and Arabie \cite{Hubert1985ComparingPartitions} concentrate on pair-counting measures for partition comparison, discussing the Rand index and introducing an innovative measure rooted in object triples. The Adjusted Rand Index (ARI), accounting for chance, enhances the Rand index and proves applicable in an iterative approach for scrutinizing clusters derived from topic modelling. Van Dongen \cite{VanDongen2000PerformanceExperiments} introduces the Markov Cluster Algorithm (MCL) tailored for graph clustering, evaluating its performance through empirical investigations. This effort culminates in a comprehensive performance criterion for clusterings of weighted graphs and a set-matching metric for comparing clusterings. In our iterative topic modelling endeavour, we harness these methods of comparison to iteratively enhance the precision of clustering results.

\section{Data Set}

The data we used to demonstrate the approach was collected from the ``COVIDSenti", a dataset from the COVID sentiment analysis project \cite{Naseem2021COVIDSenti:Analysis}. This dataset is a collection of 90,000 tweets gathered between February and March 2020. It is structured into three equal-sized subsets, each containing 30,000 tweets. Also, the tweets are categorized into three sentiment classes: positive, negative, or neutral. The labels assigned to each tweet help define its sentiment. For this project, we randomly selected a stratified sample from the first segment of 30,000 tweets with 1,000 selected from each of the sentiment classes. 

The following steps were followed to clean the raw data file before it was used for topic modelling:

\begin{itemize}
   
\item remove emojis from a given string 
\item remove URLs from a given string 
\item remove text patterns that start with a hyphen
\item remove punctuation from a given string	
\item remove user mentions, emojis, URLs, signatures, irregular patterns, and extra spacing
\item convert the message to lowercase
\item If the cleaned message is less than 15 characters or is not in English, the record was removed from further analysis

\end{itemize}
 
As mentioned, after the cleaning process, we created a subset from one of the datasets through a random selection process. Specifically, we randomly chose 3,000 tweets from the “COVIDSenti-A” dataset with an equal distribution of 1,000 tweets from each sentiment class (positive, negative, and neutral). Note that the COVIDSenti-A dataset, from which the subset is derived, is publicly accessible and widely employed in the research community for diverse studies pertaining to sentiment analysis during the COVID-19 pandemic. Note also that the data set does not have pre-defined or pre-determined label of topics and therefore its use for demonstrating our proposed approach is realistic and meaningful.

\section{The Topic Modelling Process}

Given a set of short texts, such as tweets or newsgroup postings, there are many ways to classify these texts into topics assuming texts addressing the same topic are semantically similar and therefore could be classified on that basis. For this research, we have chosen the approach used in the BERTopic package \cite{Grootendorst2022BERTopic:Procedure} and in Asyaky and Mandala \cite{Asyaky2021ImprovingUMAP}:

\begin{enumerate}
    \item Clean the short texts using a natural language processing tool kit
    \item Use the BERT embedding to convert the text data into vector representations
    \item Use the UMAP Process \cite{Mcinnes2018UMAP:Reduction} to reduce the dimensions of the data
    \item Use a hierarchical clustering algorithm HDBSCAN \cite{Campello2013Density-basedEstimates, Stewart2022AnAlgorithm} to cluster the data with reduced dimensions
\end{enumerate}

The library for BERT as well as the implementation of  UMAP, and HDBSCAN are available in Python. Therefore, the implementation of this process is straightforward.

When implementing the above process, there are several hyperparameters within UMAP and HDBSCAN that one can tune to optimize the dimension reduction and clustering process. Three parameters, namely the number of dimensions in UMAP, the cluster selection method (EOM or LEAF), and the number of clusters are the critical ones to consider. As usual, the tuning of these parameters is completely dependent on the underlying data set and the problem at hand. Further details about this process will be described in the Results section below.

Alternatively, we can use the BERTopic package that was developed by Grootendorst \cite{Grootendorst2022BERTopic:Procedure}. As its default feature, this package follows essentially the same steps as described here with additional features in the output that are useful. Therefore, we will use this BERTopic package as the topic modelling process in this research.

\section{The Iteration Steps}

The novel process proposed and studied in this research is to conduct the topic modelling process iteratively until successive clustering of the text data set produces minimum ``improvements". Conceptually, the iterative process is intuitive and is used in many situations. Details of the process as applied in this situation would require a detailed explanation.

Under the Bertopic process, an ``outlier" group, labeled ``-1" is produced in the output. This group consists of texts that have low specificity toward any cohesive topic that has been generated in the process. Therefore, based on this concept, an outline of our approach would be to:

\begin{enumerate}
  \item Choose an initial number of topics, N, we wish to create. Use the Bertopic process to segment the data set into N topics. In general, as described in the next section, efforts will be made to fine-tune the results. We call this the iteration 0 step.
  \item Put the ``-1" group of texts aside. This ``-1" group of text will be considered one topic that will not change in successive iterations. Use the Bertopic process to segment the dataset without the ``-1'' group of texts into N-k topics. k could be 1 if we see that we are close to the end of the iteration process (the indices described below such as the Adjusted Rand Index have not changed drastically). This is the iteration 1 step.  There will be another ``-1" group generated. 
  \item Calculate the Adjusted Rand index \cite{Hubert1985ComparingPartitions}, the van Dongen measure \cite{VanDongen2000PerformanceExperiments}, and the normalized variation of information index \cite{Meila2007ComparingDistance} to compare the two sets of topics generated by the successive iterations. These indices are chosen to measure how similar the clusterings produced in the two iterations are \cite{Meila2007ComparingDistance}. Calculating these three indices using the ``mclustcomp" package in R \cite{You2022Packagemclustcomp} is straightforward. 
  \item Repeat Steps 2 to 3 until the difference in the index values for successive iterations is smaller than a desirable number, say, 0.02. This is the stopping rule. Note that in practice, using one of the three indices in the stopping rule will be sufficient.
\end{enumerate}

The final set of topics would be the collection of ``-1" groups plus the topics from the last topic modelling process. 

\section{Results}

In this section, we report the results of the topic modelling process using the iterative approach using the selected data set. 

Step 0: In the initial step. We did not set a specific number of topics. Instead, we created the ``default" model using the BERTtopic package. This resulted in a clustering of 52 topics. In addition, 1,172 tweets were assigned to the ``-1" group (outliers). 

Next, four iterative steps as outlined above were carried out. The following are the details of each iteration:

Step 1: This iteration reflects the results from putting the ~-1" group of tweets from Step 0 aside and reducing the topics from the original model by 1 (N-1) and creating the model again specifying 51 as the number of clusters. Consequently, the model produced a new topic distribution with 48 clusters, including a new ``-1" group.

Step 2: Based on the results obtained in Iteration 1, we once again removed outliers (topic -1) from the dataset. We performed the topic modelling process, specifying 47 as the number of clusters, resulting from reducing the topics from the first iteration by 1 (N-1). Ultimately, the model provided a new topic distribution with 38 clusters (including -1 and 0).

Step 3 (37 Topics): The process was repeated, utilizing the results from Iteration 2. Outliers were removed, and 37 was specified as the number of topics, resulting from reducing the topics from the second iteration by 1 (N-1). This time, we indeed obtained 37 as the number of topics (including -1 and 0).

Step 4 (36 topics): In this final iteration, building upon Iteration 3, we once again removed the outlier and conducted the modelling process for the fourth time, specifying 36 as the number of clusters. The result of this ultimate iteration was a new topic distribution with 36 topics, including -1 and 0.

A file was then created with results from all iterations.

The following table summarizes detail of the five iterative steps:

\begin{table} [!ht]
    \centering
    \caption{Details of the Five Iterations}
    \begin{tabular}{|l|l|l|}
    \hline
        Iteration & Number of Topics & Size of Topic -1 \\ \hline
        0 & 52 & 1,172 \\ \hline
        1 & 48 & 493 \\ \hline
        2 & 38 & 103 \\ \hline
        3 & 37 & 84 \\ \hline
        4 & 36 & 48 \\ \hline
    \end{tabular}
\end{table}

 Additionally, we generated four cluster comparison matrices. Each matrix compared the topics generated by one iteration with the subsequent iteration. These matrices revealed that reducing the number of topics contributes to increased cluster stability.

The results of the comparisons are as follows:

\begin{table}[!ht]
   \centering
    \caption{Changes in successive iterations of the Three Clustering Comparison Indices}
    \begin{tabular}{|l|l|l|l|}
    \hline
        Iteration & Adjusted Rand & Van Dongen & NVI\\ \hline
        0 vs 1 & 0.83 & 0.17 & 0.34 \\ \hline
        1 vs 2 & 0.83 & 0.06 & 0.12 \\ \hline
        2 vs 3 & 0.91 & 0.05 & 0.08 \\ \hline
        3 vs 4 & 0.98 & 0.02 & 0.04 \\ \hline
    \end{tabular}
\end{table}

Note that the Adjusted Rand index (ARI) is very close to 1. On the other hand, both the Van Dongen and the normalized variation of information (NVI) metrics are close to 0, indicating that  clustering results of iterations 3 and 4 are very close. It is a signal that we should stop the iteration process and use the results from iteration 4 as the result of the topic modelling exercise.

\section{Discussions}


Evaluating the ``accuracy" of a cluster analysis in general and a topic modelling exercise in particular is very difficult if not impossible unless the ``ground" truth of the clustering is available. The iterative approach used here gives at least the comfort that no further improvements need to be made using the topic modelling method of choice.

The plan for further research is therefore clear. The iterative approach should be applied to a data set with ``ground truth" on the topic and see how it would perform. The data set documented in Miles et al \cite{Miles2022AModeling} would be a good candidate. The other direction of research would be the adaption of the iterative approach to other topic modelling algorithms.

\section{Conclusions}
In this paper, we have delved into the latest advance in topic modelling techniques, highlighting the integration of algorithms such as BERT, UMAP, clustering algorithms, and an iterative approach. This advancement represents a leap forward in the realm of text analysis, as they synergistically enhance clustering algorithms, resulting in heightened levels of precision and efficacy.

Our study presents a framework that not only encompasses aspects of text processing but also extends to the nuanced intricacies of topic modelling. Through experimentation and evaluation, we have crafted a workflow tailored toward transforming raw textual data into structured insights. Leveraging the publicly available Twitter dataset as our testing ground, we embarked on the development and assessment of an iterative approach within the domain of topic modelling.

One of the outcomes of our research is the tangible reduction in the number of topics, from an initial 52 down to 36 topics, following four iterative cycles. This process demonstrates the efficacy of our approach in refining the clustering process, thereby facilitating more concise topic delineation. Moreover, using a measure of clustering comparison, these topics are deemed to be complete in the sense that no significant improvements could be made using the modelling approach.

We believe that our exploration merely scratches the surface of the potential of this approach. The early success leads us to believe that further research in using this approach in conjunction with other topic modelling algorithms could be viable.

\bibliographystyle{IEEEtran}
\balance
\bibliography{TopicModel}

\end{document}